\documentclass[sigconf]{acmart}

\renewcommand\footnotetextcopyrightpermission[1]{}

\AtBeginDocument{%
  \providecommand\BibTeX{{%
    \normalfont B\kern-0.5em{\scshape i\kern-0.25em b}\kern-0.8em\TeX}}}

\setcopyright{acmlicensed}
\copyrightyear{2018}
\acmYear{2018}
\acmDOI{XXXXXXX.XXXXXXX}

\acmConference[MM'24]{Make sure to enter the correct
  conference title from your rights confirmation email}{October 28 - November 1,
  2024}{Melbourne, Australia.}
\acmISBN{978-1-4503-XXXX-X/18/06}

\begin{document}

\title{MambaDFuse: A Mamba-based Dual-phase Model for Multi-modality Image Fusion}


\author{Zhe Li}
\email{lizhe99@hrbeu.edu.cn}
\affiliation{%
  \institution{Harbin Engineering University}
  \city{}
  \country{}}

\author{Haiwei Pan}
\authornote{Corresponding authors.}
\email{panhaiwei@hrbeu.edu.cn}
\affiliation{%
  \institution{Harbin Engineering University}
  \city{}
  \country{}}

\author{Kejia Zhang}
\email{kejiazhang@hrbeu.edu.cn}
\affiliation{%
  \institution{Harbin Engineering University}
  \city{}
  \country{}}

\author{Yuhua Wang}
\email{wangyuhua@hrbeu.edu.cn.}
\affiliation{%
	\institution{Harbin Engineering University}
	\city{}
	\country{}
}

\author{Fengming Yu}
\email{yufengming@hrbeu.edu.cn}
\affiliation{%
	\institution{Harbin Engineering University}
	\city{}
	\country{}}
\renewcommand{\shortauthors}{author name and author name, et al.}

\begin{abstract}
Multi-modality image fusion (MMIF) aims to integrate complementary information from different modalities into a single fused image to represent the imaging scene and facilitate downstream visual tasks comprehensively. In recent years, significant progress has been made in MMIF tasks due to advances in deep neural networks. However, existing methods cannot effectively and efficiently extract modality-specific and modality-fused features constrained by the inherent local reductive bias (CNN) or quadratic computational complexity (Transformers). To overcome this issue, we propose a Mamba-based Dual-phase Fusion (MambaDFuse) model. Firstly, a dual-level feature extractor is designed to capture long-range features from single-modality images by extracting low and high-level features from CNN and Mamba blocks. Then, a dual-phase feature fusion module is proposed to obtain fusion features that combine complementary information from different modalities. It uses the channel exchange method for shallow fusion and the enhanced Multi-modal Mamba (M3) blocks for deep fusion. Finally, the fused image reconstruction module utilizes the inverse transformation of the feature extraction to generate the fused result. Through extensive experiments, our approach achieves promising fusion results in infrared-visible image fusion and medical image fusion. Additionally, in a unified benchmark, MambaDFuse has also demonstrated improved performance in downstream tasks such as object detection. Code with checkpoints will be available after the peer-review process.
\end{abstract}


\keywords{Image fusion, Mamba, multi-modality images, cross-modality interaction}



\maketitle
\begin{figure*}
  \includegraphics[width=\textwidth]{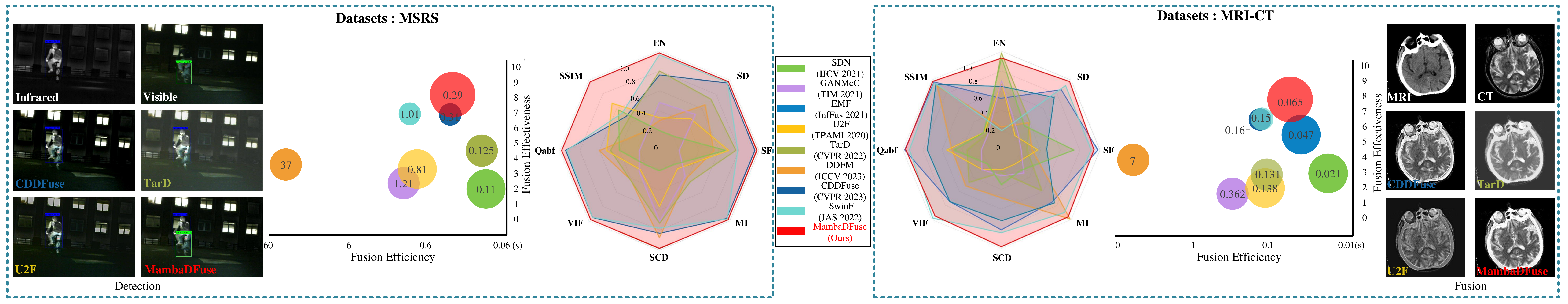}
  \caption{Fusion, detection and efficiency\&effectiveness comparisons with state-of-the-art methods on MSRS and MRI-CT datasets. Octagons formed by lines of different colors represent the values of different methods across eight metrics. Our MambaDFuse outperforms the most comprehensive performance. The bubble chart illustrates the comparative analysis of efficiency and effectiveness, and the numbers inside the circles represent the time required to fuse a pair of images. Methods achieving similar fusion performance to ours demonstrate a slower fusion rate. Conversely, methods with a slightly faster rate than ours exhibit significantly lower fusion effectiveness. The fusion and detection results also showcase the powerful fusion capabilities of MambaDFuse. (The fusion metrics used in the chart are computed after normalization. The horizontal axis of the bubble chart represents time, while the vertical axis represents the sum of metrics.)}
  \label{fig:intro}
\end{figure*}

\section{Introduction}
Image fusion aims to combine essential information representations from multiple source images to generate high-quality, content-enriched fused images \cite{ma2017multi, xu2021stereo, yang2022sir, yao2019spectral, zhao2020bayesian}. Depending on the differences in imaging devices or imaging settings, image fusion can be categorized into various types, including multi-modality image fusion (MMIF) \cite{liu2021learning, xu2020u2fusion, zhao2020didfuse}, digital photographic image fusion \cite{ma2017multi, zhang2021deep}, and remote sensing image fusion \cite{bandara2022hypertransformer, xu2021deep, zhao2021fgf, he2024pan}. Infrared-visible image fusion (IVF) and medical image fusion (MIF) are two typical tasks of MMIF, which model and fuse cross-modal features from all sensors. In particular, the infrared sensor captures thermal radiation data, highlighting prominent targets, while the visible sensor captures reflected light information, producing digital images rich in texture details \cite{ma2021stdfusionnet}. IVF aims to integrate complementary information from source images, resulting in high-contrast fusion images highlighting prominent targets while preserving rich texture details. These fusion images offer enhanced scene representation and visual perception, facilitating subsequent practical visual applications such as multi-modal saliency detection \cite{liu2019perceptual, qin2019basnet, wang2021dual}, object detection \cite{bochkovskiy2020yolov4, liu2021training, tang2021robustart}, and semantic segmentation \cite{liu2021searching, qin2022bibert, qin2020forward, qin2023distribution}. Similarly, in medical imaging, structural images such as computed tomography (CT) and magnetic resonance imaging (MRI) primarily provide structural and anatomical information \cite{awad2019adaptive}, while functional images such as positron emission tomography (PET) and single-photon emission computed tomography (SPECT) reflect the metabolic activity of normal and pathological tissues and the cerebral blood flow signals \cite{xu2021emfusion}. MIF can precisely detect abnormal locations by integrating multiple imaging modalities, thereby assisting in diagnosis and treatment \cite{he2023hqg, james2014medical}.

In recent years, numerous methods have been developed to address challenges in MMIF. These methods primarily include Convolutional Neural Networks (CNNs) and AutoEncoders (AEs) \cite{li2021rfn, li2018densefuse, liang2022fusion, zhao2020didfuse}, Generative Adversarial Networks (GANs) \cite{liu2022target, ma2020ddcgan, ma2019fusiongan, ma2020ganmcc}, Diffusion models \cite{zhao2023ddfm, yue2023dif}, as well as those based on Transformers \cite{ma2022swinfusion, qu2022transmef, zhao2023cddfuse}. The main drawback of the paradigms above is their inability to achieve both efficiency and effectiveness in MMIF. Firstly, CNN-based methods struggle to capture global context due to their limited receptive fields, making it challenging to generate high-quality fusion images. Moreover, using AE for feature extraction or image reconstruction poses challenges in designing an encoder that can capture modality-specific and shared features. Secondly, a pipeline based on generative models can generate high-quality fusion images but may not efficiently complete fusion tasks. GANs suffer from unstable training and mode collapse, while diffusion models face challenges such as long training time and slow sampling rate. Finally, models based on Transformers or a combination of Transformers and others exhibit excellent performance in global modeling but suffer from significant computational overhead due to the quadratic growth of resources with the number of token stemming from the self-attention mechanism.

The emergence of the improved S4, also known as Mamba \cite{gu2023mamba}, with its selective mechanism and efficient hardware-aware design, provides a novel solution to the challenges above. Mamba has been demonstrated to outperform Transformers in tasks requiring long-term dependency modeling, such as natural language processing, due to its input-adaptive and global information modeling capabilities while maintaining linear complexity, reducing computational costs, and enhancing inference speed. Recently, some variants of Mamba have also shown promising results in computer vision tasks, such as image classification \cite{zhu2024vision, liu2024vmamba}, medical image segmentation \cite{ma2024u}, and so on. However, Mamba's role in the MMIF task has yet to be fully explored, as Mamba lacks a design similar to cross-attention. This prompts us to investigate how to utilize Mamba to integrate multi-modal information in MMIF.

Therefore, we propose a \textbf{Mamba}-based \textbf{D}ual-phase Model for Multi-modality Image \textbf{Fusion}(\textbf{MambaDFuse}). It consists of three stages: a dual-level feature extraction, a dual-phase feature fusion, and fused image reconstruction. The hierarchical feature extraction consists of convolutional layers and multiple stacked Mamba blocks, leveraging the excellent processing capabilities of CNNs in the early stages of visual tasks and the efficiency of Mamba in extracting long-range features. Then, in feature fusion, the shallow fuse module utilizes manually designed fusion rules to fuse global overview features. In contrast, the deep fuse module conducts cross-modal deep feature fusion with improved Multi-modal Mamba (M3) blocks to obtain local detail features guided by the respective modality features. Ultimately, the fusion features are used to reconstruct the fused image. The loss function for reconstructing adopts the losses proposed in \cite{ma2022swinfusion}, consisting of SSIM loss, texture loss, and intensity loss, driving the network to preserve rich texture details and structural information while presenting optimal visual intensity. Fig. \ref{fig:intro} demonstrates that our proposed MambaDFuse outperforms state-of-the-art methods in terms of subjective visual assessment and objective evaluation metrics.

Our main contributions can be summarized as follows:
\begin{itemize}
\item To the best of our knowledge, MambaDFuse is the first to leverage Mamba for MMIF, which is an alternative to CNNs and Transformers with effectiveness and efficiency.
\item To capture low and high-level modality-specific features with long-range information, we design a dual-level feature extractor. The features encompass prominent objects, environmental lighting, and texture details.
\item To get modality-fused features with global overview and local detail information, we propose a dual-phase feature fusion module. Specifically, channel exchange is used for shallow fuse, and an M3 block is designed for deep fuse.
\item Our method achieves leading image fusion performance for both IVF and MIF. We also present a unified measurement benchmark to justify how the IVF fusion images facilitate downstream object detection.
\end{itemize}


\section{Related Works}
\subsection{Multi-Modality image fusion}
Multi-modality image fusion research based on deep learning leverages the powerful fitting capability of neural networks, enabling effective feature extraction and information fusion. According to the backbone of the models, existing methods can be classified into three categories: methods based on CNNs and AEs, methods employing generative models such as GANs and Diffusions, and methods utilizing Transformer (some in conjunction with CNN).

Firstly, for methods based on CNN and AE \cite{li2021rfn, li2018densefuse, zhao2020didfuse, li2020nestfuse, liu2021searching, zhang2021sdnet}, the typical pipeline involves feature extraction using CNNs or encoders, followed by image reconstruction using AEs. Utilizing context-agnostic CNNs is their limitation in extracting global information to generate high-quality fusion images, leading to a solid local reductive bias. Therefore, whether CNNs are sufficient for extracting features from all modalities remains to be seen. Additionally, careful consideration is required for encoder design. Shared encoders may fail to distinguish modality-specific features, while using separate encoders may overlook shared features between modalities.

Secondly, for methods based on generative models, GANs \cite{liu2022target, ma2020ddcgan, ma2019fusiongan, ma2020ganmcc} model the image fusion problem as an adversarial game between a generator and a discriminator, using adversarial training to generate fusion images with the same distribution as the source images. However, unstable training, lack of interpretability, and mode collapse have been critical issues affecting the generative capabilities of GANs. Also, diffusion models, consisting of diffusion and inverse diffusion stages, have achieved remarkable success in image generation. However, when applied to the task of multi-modal image fusion \cite{zhao2023ddfm, yue2023dif}, an unresolved challenge is the high computational cost, resulting in long training and inference time, which impacts the efficiency of the fusion process.

Finally, methods based on Transformer or CNN-Transformer \cite{ma2022swinfusion, qu2022transmef, zhao2023cddfuse} have shown promising results due to their strong ability to model long-range dependencies. Transformers and their variants are used for feature extraction, fusion, and image reconstruction. However, the quadratic time complexity and computational resource consumption associated with the self-attention mechanism make them inefficient for multi-modal image fusion tasks. Even methods like [40], which use shifted window attention to improve performance, compromise long-term dependencies and fundamentally fail to address the quadratic complexity. 

In conclusion, image fusion, a preprocessing step for advanced visual tasks, requires real-time processing and robust fusion capabilities. However, existing approaches have yet to achieve effective and efficient image fusion performance. Therefore, there is an urgent need for a new backbone to propel the progress and development of MMIF.

\begin{figure*}
  \centering
  \includegraphics[width=\textwidth]{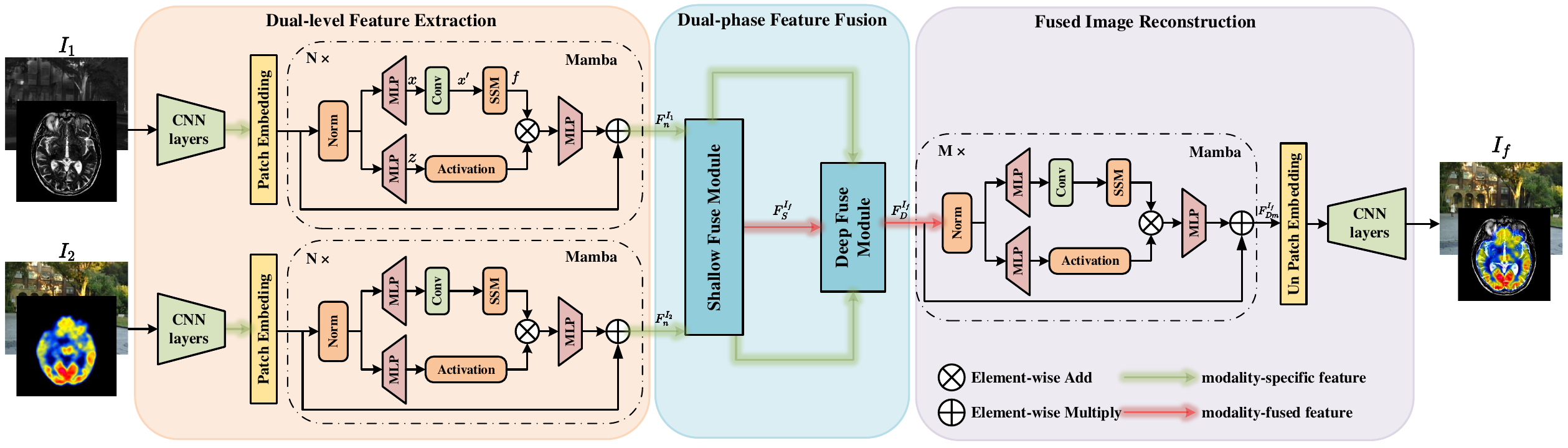}
  \caption{The overall architecture of MambaDFuse. It consists of three stages: dual-level feature extraction, dual-phase feature fusion, and fused image reconstruction.}
  \label{fig:overall}
\end{figure*}

\subsection{State Space Models (SSMs)}
State Space Models (SSM) \cite{gu2021efficiently, gu2021combining, smith2022simplified}, originating from classical control theory \cite{kalman1960new}, have become practical components for constructing deep networks due to their cutting-edge performance in analyzing continuous long sequential data. Structured State Space Sequence Models (S4) \cite{gu2021efficiently} represent pioneering work in state space models for modeling long-range dependencies. Subsequently, the S5 layer \cite{smith2022simplified} was proposed based on S4, introducing a parallel scan on a diagonalized linear SSM. The H3 model \cite{fu2022hungry} further refined and extended this work, enabling the model to achieve results comparable to Transformer in language modeling tasks.

A recent study named Mamba \cite{gu2023mamba} further improved upon S4 by introducing a selection mechanism, enabling the model to choose relevant information depending on the input selectively. Simultaneously, a hardware-aware algorithm was proposed to achieve efficient training and inference. Compared to Transformer models of equivalent scale, Mamba exhibits higher inference speed, throughput, and overall performance. Subsequently, many works have extended Mamba from natural language processing (NLP) to other domains \cite{wang2023selective, nguyen2022s4nd, islam2023efficient}. Visual Mamba (Vim) \cite{zhu2024vision} applies Mamba to the Vision Transformer (ViT) architecture, proposing a novel universal visual backbone based on bidirectional Mamba blocks. This backbone embeds positional embeddings into image sequences and compresses visual representations through bidirectional state space models. Visual State Space Model (Vamba) \cite{liu2024vmamba} introduces a cross-scanning mechanism to bridge the gap between one-dimensional array scanning and two-dimensional plane traversal. In medical image segmentation tasks, Mamba has also been applied \cite{ma2024u}, yielding promising results. Since Mamba lacks a mechanism like cross-attention that facilitates the fusion of multi-modal information, research on effectively utilizing Mamba in MMIF is worthy of study.

In conclusion, similar to works such as \cite{ma2022swinfusion}, MambaDFuse comprises feature extraction, feature fusion, and image reconstruction. However, what sets our approach apart is the utilization and enhancement of Mamba Blocks. Specifically, we design a dual-level feature extractor, an M3 block and a dual-phase feature fusion module tailored for MMIF. And through such a design, MambaDFuse can become a powerful tool in MMIF.

\section{Method}
\subsection{Preliminaries}
State space models(SSMs) are conventionally considered as a linear time-invariant system that builds a map stimulation from $x\left ( t \right ) \in \mathbb{R}^{N}$ to response $y\left ( t \right ) \in \mathbb{R}^{N}$ by a latent state $y\left ( h \right ) \in \mathbb{R}^{N}$. The system can be mathematically expressed using a linear ordinary differential equation (ODE):
\begin{align}
 h'\left ( t \right ) &= Ah\left ( t \right ) + Bx\left ( t \right ) \label{con:1}\\
y\left ( t \right ) &= Ch\left ( t \right )\label{con:2}
\end{align}
where $N$ is the state size, $A\in \mathbb{R}^{N\times N}$, $B\in \mathbb{R}^{N\times 1}$, and $C\in \mathbb{R}^{N\times 1}$.

As continuous-time models, SSMs pose significant challenges when incorporated into deep learning algorithms. To surmount this barrier, discretization becomes indispensable. We can discretize the system in Eq. \ref{con:1} using the method of zero-order hold (ZOH), which can be formally defined as follows:
\begin{align}
h_{t}&=\overline{A}h_{t-1} +  \overline{B}x_{t} \label{con:3}\\
y_{t}&=Ch_{t}\label{con:4}
\end{align}
where $\overline{A}=\exp \left ( \Delta A \right )  $ and $\overline{B}=\left ( \Delta A \right )^{-1} \left ( \exp\left (  \Delta A\right ) -I  \right ) \cdot \Delta B$ are the discretized state parameters and $\Delta$ is the discretization step size.

However, the expression presented in Eq. \ref{con:3} and \ref{con:4} concentrates on the Linear Time-Invariant System with parameters that remain constant for different inputs. To overcome this constraint, Mamba \cite{gu2023mamba} has sought to integrate a selective scanning mechanism, wherein the matrices $B$, $C$, and $\Delta$ are derived from the input data. In addition, by relying on a faster hardware-aware algorithm, Mamba has further advanced its potential.

\subsection{Overall Architecture}
The proposed MambaDFuse can be divided into three functional components: a dual-level feature extractor, a dual-phase feature fusion module, and a fused image reconstruction module. The detailed workflow is illustrated in Fig. \ref{fig:overall}. Assuming two different modalities of images, $I_{1}$ and $I_{2}$, with $I_{f}$ representing the fusion image, our pipeline can be described as follows:

Initially, dual-level feature extraction consists of low-level and high-level feature extraction. In the former, $I_{1}$ and $I_{2}$ are projected into a shared feature space by convolutional layers $\varphi$. However, the CNN layers may fail to capture global features due to their limited receptive fields. Therefore, after patch embedding, N $\times $ Mamba blocks $\phi$ are utilized for high-level feature extraction,  yielding modality-specific features $ F_{n}^{I_{1}} $ and $ F_{n}^{I_{2}} $.
\begin{align}
F_{0}^{I_{1}}, F_{0}^{I_{2}} &= PatchEmbed\left ( \varphi\left ( I_{1}  \right )   \right ),PatchEmbed\left ( \varphi\left ( I_{2}  \right )   \right ) \label{con:5}\\
F_{n}^{I_{1}}&= \phi _{1n}  \cdot \cdot \cdot  \left ( \phi _{11} \left (  \phi _{10} \left ( F_{0}^{I_{1}}  \right ) \right )     \right )\label{con:6}\\
F_{n}^{I_{2}}&= \phi _{2n}  \cdot \cdot \cdot  \left ( \phi _{21} \left (  \phi _{20} \left ( F_{0}^{I_{2}}  \right ) \right )     \right )\label{con:7}
\end{align}

Subsequently, in the shallow fuse module, a manually designed fusion strategy is employed to obtain an initial fusion feature $F_{S}^{I_{f}}$, including global integrated information. As for the local detail feature, we design an Multi-modal Mamba (M3) block to construct a deep fuse module, which utilizes multi-modal features as guidance to derive modality-fused feature $F_{D}^{I_{f}}$ for reconstruction.
\begin{align}
 F_{n}^{I_{1}'},F_{n}^{I_{2}'},F_{S}^{I_{f}}&=ShallowFuse\left ( F_{n}^{I_{1}},F_{n}^{I_{2}} \right ) \label{con:8}\\ 
F_{D}^{I_{f}}&=DeeepFuse\left ( F_{n}^{I_{1}'},F_{n}^{I_{2}'},F_{S}^{I_{f}} \right ) \label{con:9}
\end{align}

Lastly, in the fused image reconstruction stage, contrary to the feature extraction, the feature $F_{D}^{I_{f}}$ undergo M $\times $ Mamba blocks $\phi$, followed by unpatchembedding, and finally, convolutional layers are applied to obtain the fused result $I_{f}$. 
\begin{align}
F_{Dm}^{I_{f}}&= \phi _{fm}  \cdot \cdot \cdot  \left ( \phi _{f1} \left (  \phi _{f0} \left ( F_{D0}^{I_{f}}  \right ) \right )     \right )\label{con:10}\\
I_{f}  &= \varphi  \left ( UnPatchEmbed\left ( F_{Dm}^{I_{f}}  \right )  \right ) \label{con:11}
\end{align}
In the following sections (\ref{dlfe} and \ref{dlff}), the detail and motivation of dual-level feature extraction and dual-phase feature fusion will be provided.  

\subsection{Dual-level Feature Extraction}\label{dlfe}

\textbf{Low-level Feature Extraction.} Convolutional layers excel in early visual processing, enhancing optimization stability and superior outcomes \cite{xiao2021early}. Additionally, they provide a straightforward yet efficient means of capturing local semantic details and projecting them into a high-dimensional feature space. The Low-level Feature Extraction comprises two convolutional layers utilizing the Leaky ReLU activation function, each with a kernel size of 3×3 and a stride of 1.\\
\begin{algorithm}
\small
    \caption{High-level Feature Extraction.}
    \label{alg:Algorithm1}
    \renewcommand{\algorithmicrequire}{\textbf{Input:}}
    \renewcommand{\algorithmicensure}{\textbf{Output:}}
    \begin{algorithmic}[1]
        \Require feature sequence $\mathbf{F_{t-1}}:\rm{\textcolor[RGB]{0, 100, 0}{(B,N,C)}}$
        \Ensure feature sequence $\mathbf{F_{t}}:\rm{\textcolor[RGB]{0, 100, 0}{(B,N,C)}}$
        \State  $\mathbf{F_{t-1}'}:\rm{\textcolor[RGB]{0, 100, 0}{(B,N,C)}} \gets \mathbf{LayerNorm(F_{t-1})}$
        \State  $\mathbf{x}:\rm{\textcolor[RGB]{0, 100, 0}{(B,N,C')}} \gets \mathbf{MLP_{x}(F_{t-1}')}$
        \State  $\mathbf{z}:\rm{\textcolor[RGB]{0, 100, 0}{(B,N,C')}} \gets \mathbf{MLP_{z}(F_{t-1}')}$
        \State  $\mathbf{x'}:\rm{\textcolor[RGB]{0, 100, 0}{(B,N,C')}} \gets \mathbf{SiLU(\mathbf{Conv}(x))}$
        \Statex \textcolor{gray}{\text{/* $\mathbf{Disc}$ and $\mathbf{SSM}$ represents Eq. \ref{con:3} and \ref{con:4} implemented by selective scan \cite{gu2023mamba} */}}
        \State  $\mathbf{\overline{A}}:\rm{\textcolor[RGB]{0, 100, 0}{(B,N,C',D)}}, \mathbf{\overline{B}}:\rm{\textcolor[RGB]{0, 100, 0}{(B,N,C',D)}}, \mathbf{C}:\rm{\textcolor[RGB]{0, 100, 0}{(B,N,C',D)}} \gets \mathbf{Disc(x')}$
        \State  $\mathbf{f}: \rm{\textcolor[RGB]{0, 100, 0}{(B,N,C')}} \gets \mathbf{SSM(\overline{A}, \overline{B}, C)(x')}$
        \Statex \textcolor{gray}{\text{/* $\mathbf{F_{t}}$ represents modality-specific features*/}}   
        \State  $\mathbf{f'}:\rm{\textcolor[RGB]{0, 100, 0}{(B,N,C')}}\gets \mathbf{f \odot SiLu(z)}$
        \State  $\mathbf{F_{t}}:\rm{\textcolor[RGB]{0, 100, 0}{(B,N,C)}} \gets \mathbf{MLP_{F}(f') + F_{t-1}}$
        \Statex $\mathbf{return}\; \mathbf{F_{t}}$
    \end{algorithmic}
\end{algorithm}
\textbf{High-level Feature Extraction.} CNNs are constrained by local receptive fields when extracting features, while Transformers also encounter the issue of quadratic complexity. Considering this, mamba blocks are a promising choice for further extracting advanced modality-specific features. An input feature sequence $F_{t-1}\in \mathbb{R}^{B\times N\times C }$, firstly is applied layer normalization to obtain $F_{t-1}'$. Then, on two separate branches, $F_{t-1}'$ is projected into $x$ and $z$ using two multi-layer perceptrons (MLP). On the first branch, $z$ undergoes convolution and SiLU activation to obtain $x'$. Subsequently, $x'$ undergoes the State Space Model (Eq. \ref{con:3} and \ref{con:4}) to compute $f$. On the other branch, $z$ passes through a SiLU activation function to serve as a gating factor for gating $f$, resulting in $f'$. Finally, after an MLP layer and a residual connection, the output of the high-level feature extraction, $F_{t}$, is obtained. The detailed process is shown in Algorithm \ref{alg:Algorithm1}.

\subsection{Dual-phase Feature Fusion}\label{dlff}
\begin{figure*}
    \centering
    \includegraphics[width=1\linewidth]{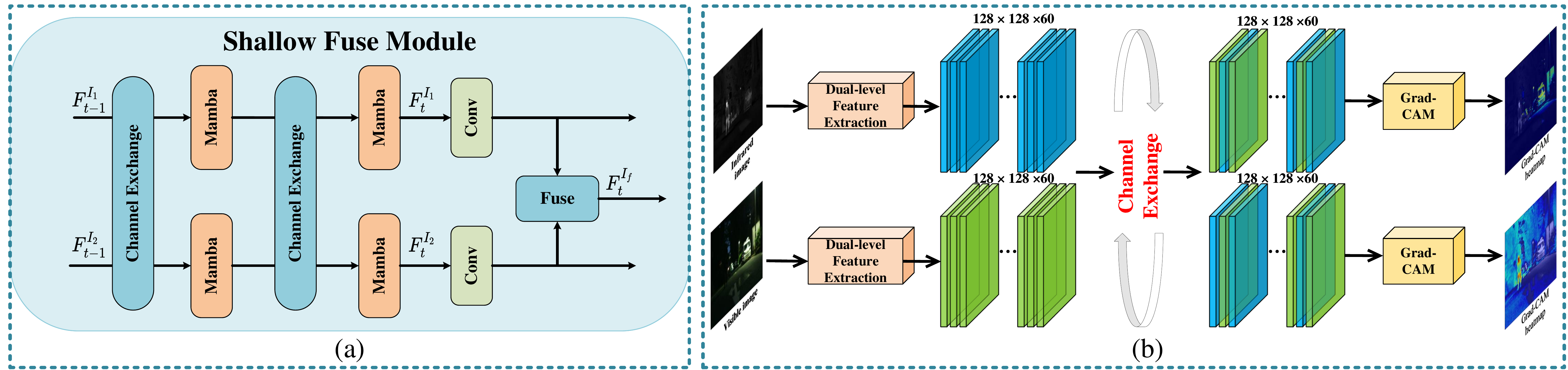}
    \caption{(a) is the implementation details of the shallow fuse module. (b) is channel exchange process and grad-cam \cite{selvaraju2017grad} visualization results. The visualization is calculated for the output of Conv in the shallow fuse module. The heatmap demonstrates that the image has integrated information from the other modality after channel exchange, contributing to a shallow fuse.}
    \label{fig:shallow}
\end{figure*}

\begin{figure}
    \centering
    \includegraphics[width=1\linewidth]{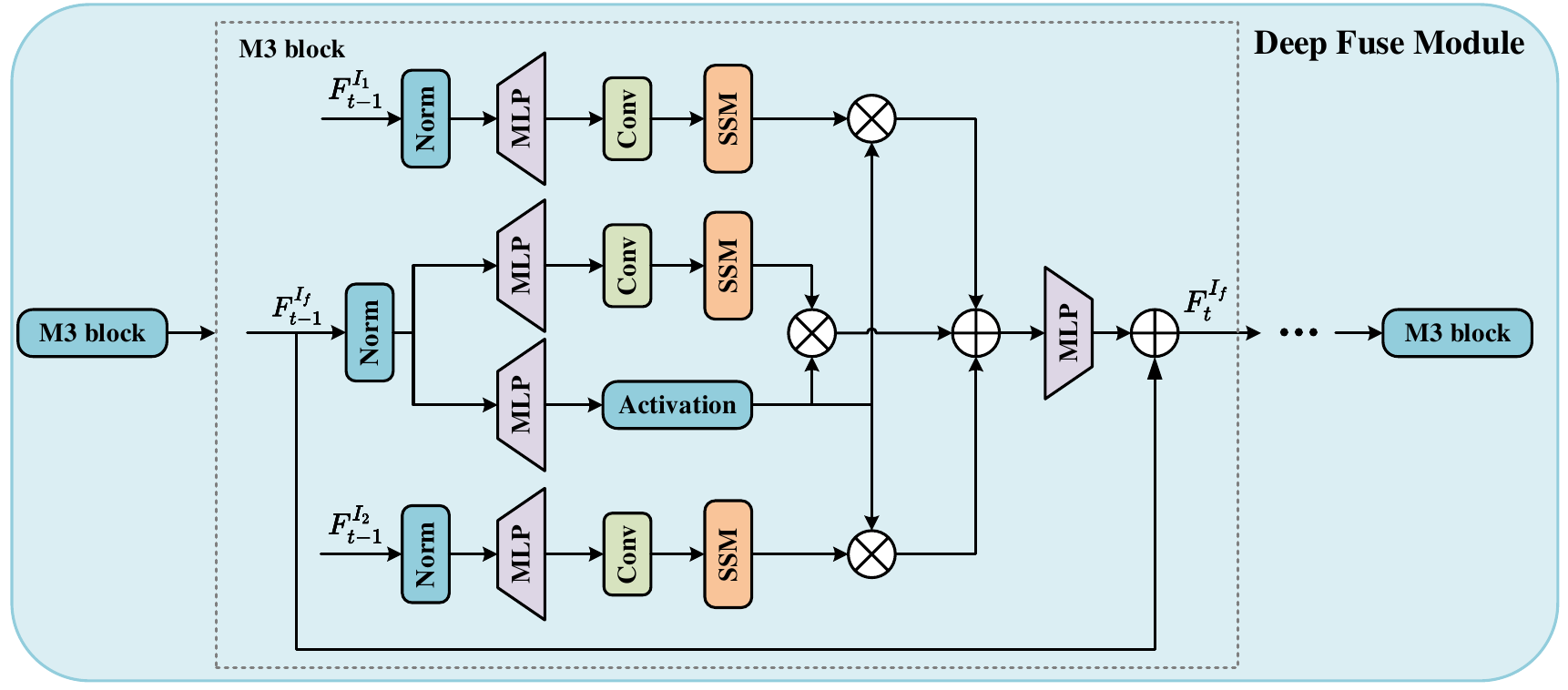}
    \caption{The implementation details of the deep fuse module.}
    \label{fig:deep}
\end{figure}

A practical fusion feature should incorporate significant information such as salient objects, environmental lighting, and texture details. A manually designed fusion rule can quickly produce an initial fusion feature in the first phase (as shown in Fig. \ref{fig:shallow} (a)). However, due to its inability to accommodate more complex fusion scenarios, an enhanced Mamba M3 block is utilized in the second phase (as shown in Fig. \ref{fig:deep}) to fuse deep texture detail features.\\
\textbf{Shallow Fuse Module.} Inspired by \cite{fang2023changer}, we firstly adopt a channel exchange approach (Eq. \ref{con:12}) on $F_{t-1}^{I_{1}}$ and $F_{t-1}^{I_{2}}$, which does not require additional parameters or computational operations, to achieve lightweight exchange of features from multiple modalities. The exchanged features then undergo processing through their respective Mamba blocks. By repeating the above steps, modality-specific features $F_{t}^{I_{1}}$ and $F_{t}^{I_{2}}$ incorporate features from another modality, as illustrated in Fig. \ref{fig:shallow} (b). Subsequently, they undergo a FUSE operation to obtain a shallow fusion feature $F_{t}^{I_{f}}$. The FUSE operation can be addition or L1 normalization as proposed in \cite{li2018densefuse}.
\begin{align}
F_{t-1}^{I_{1}/I_{2}}= 
\begin{cases}
    F_{t-1}^{I_{1}/I_{2}} & \text{if } M\left ( B, N,C \right )  = 0 \\
    F_{t-1}^{I_{2}/I_{1}} & \text{if } M\left ( B, N,C \right )  = 1\label{con:12}
\end{cases} 
\end{align}
$M(B,N,C)$ refers to the mask used for channel exchange, consisting of 1s and 0s where 0 indicates no exchange and 1 indicates exchange.\\
\textbf{Deep Fuse Module.} The current Mamba architecture cannot directly handle multi-modal image information because it lacks mechanisms similar to cross-attention. As an improvement, we design an Multi-modal Mamba (M3) block. It utilizes modality-specific features to guide the generation of modality-fused features, aiming to incorporate local detail characteristics from different modalities inspired by \cite{he2024pan}. The input is the initial fusion feature obtained from the shallow fuse module. Moreover, two additional branches are introduced, each taking features from different modalities as input. Similarly, these branches undergo layer normalization, convolution, SiLU activation, and parameter discretization and pass through the SSM to obtain an output $y$. After modulation by the gating factor, it is added to the output of the original branch, resulting in a final fusion feature. Please refer to the Algorithm \ref{alg:Algorithm 4} for specific details. A series of M3 blocks comprises the deep fuse module.

\begin{algorithm}\small
    \caption{ Multi-modal Mamba (M3) Block.}
    \label{alg:Algorithm 4}
    \renewcommand{\algorithmicrequire}{\textbf{Input:}}
    \renewcommand{\algorithmicensure}{\textbf{Output:}}    
    \begin{algorithmic}[1] 
        \Require feature sequence $\mathbf{F_{t-1}^{I_{1}} }:\rm{\textcolor[RGB]{0, 100, 0}{(B,N,C)}} , \mathbf{F_{t-1}^{I_{2}} }: \rm{\textcolor[RGB]{0, 100, 0}{(B,N,C)}} , \mathbf{F_{t-1}^{I_{f}} }:\rm{\textcolor[RGB]{0, 100, 0}{(B,N,C)}} $
        \Ensure feature sequence $\mathbf{F_{t}^{I_{f}}}:\rm{\textcolor[RGB]{0, 100, 0}{(B,N,C)}}$   
        \For{$ \mathbf{o} \ \mathbf{in} \mathbf{\left \{ I_{1},I_{2},I_{f}  \right \}} $}
            \State $\mathbf{F_{t-1}^{o'}}:\rm{\textcolor[RGB]{0, 100, 0}{(B,N,C)}}\gets \mathbf{LayerNorm(F_{t-1}^{o})}$
            \State $\mathbf{x_{o}}:\rm{\textcolor[RGB]{0, 100, 0}{(B,N,C')}} \gets \mathbf{MLP_{x_{o}}(F_{t-1}^{o'})}$
            \State $\mathbf{x_{o}'}:\rm{\textcolor[RGB]{0, 100, 0}{(B,N,C')}} \gets \mathbf{SiLu({Conv}_{o}(x_{o}))}$
            \Statex \textcolor{gray}{\text{/* $\mathbf{Disc}$ and $\mathbf{SSM}$ represents Eq. \ref{con:3} and \ref{con:4} implemented by selective scan \cite{gu2023mamba} */}}
            \State $\mathbf{\overline{A}}:\rm{\textcolor[RGB]{0, 100, 0}{(B,N,C',D)}}, \mathbf{\overline{B}}:\rm{\textcolor[RGB]{0, 100, 0}{(B,N,C',D)}},\mathbf{C}:\rm{\textcolor[RGB]{0, 100, 0}{(B,N,C',D)}}\gets \mathbf{Disc(x_{o}')}$
            \State $\mathbf{y_{o}}:\rm{\textcolor[RGB]{0, 100, 0}{(B,N,C')}} \gets \mathbf{SSM(\overline{A}, \overline{B}, C)(x_{o}')}$ 
        \EndFor
        \State $\mathbf{z}:\rm{\textcolor[RGB]{0, 100, 0}{(B,N,C')}} \gets \mathbf{MLP_{z}(F_{t-1}^{If'})}$
        \Statex \textcolor{gray}{\text{/* $\mathbf{F_{t}^{I_{f}}}$ represents modality-fused features guided by modality-specific features*/}}
        \State $\mathbf{y_{I_{1}}'}:\rm{\textcolor[RGB]{0, 100, 0}{(B,N,C')}} \gets \mathbf{y_{I_{1}} \odot {SiLu}(z)}$
        \State $\mathbf{y_{I_{2}}'}:\rm{\textcolor[RGB]{0, 100, 0}{(B,N,C')}} \gets \mathbf{y_{I_{2}} \odot {SiLu}(z)}$
        \State $\mathbf{F_{t}^{I_{f}}}:\rm{\textcolor[RGB]{0, 100, 0}{(B,N,C)}} \gets \mathbf{MLP_{F}(y_{I_{1}}'+y_{I_{2}}')+F_{t-1}^{I_{f}}}$
        \Statex  $\mathbf{return}\;\mathbf{F_{t}^{I_{f} }}$; 
    \end{algorithmic}
\end{algorithm}

\section{Experiments}
\subsection{Setup}
\textbf{Datasets.} For IVF experiments, three datasets are used to verify MambaDFuse, i.e., MSRS \cite{tang2022piafusion}, RoadScene \cite{xu2020fusiondn}, and $\mathrm{M^{3}FD}$ \cite{liu2022target}. The training dataset consists of an MSRS training set (1083 pairs), 200 pairs in $\mathrm{M^{3}FD}$ and 151 pairs in RoadScene, while an MSRS testing set (361 pairs), 100 pairs in $\mathrm{M^{3}FD}$ and 70 pairs in RoadScene are employed as the testing dataset, which the fusion performance can be verified comprehensively. The datasets contain a diverse range of images captured night and day. Most images are taken on roads and include objects such as people, cars, bicycles, and road signs. For MIF experiments, we selected pairs of medical images from the Harvard Medical website \cite{Harvardmedicalwebsite}, of which 48 pairs of MRI-CT images, 190 pairs of MRI-PET images, and 81 pairs of MRI-SPECT images.\\
\begin{table}\small
  \caption{Quantitative results of the IVF task. \bfseries\textcolor{red}{Red} and \bfseries\textcolor{blue}{blue} show the best and second-best values, respectively.}
  \label{tab:VIF}
  \begin{tabular}{ccccccccc}
    \toprule
    \multicolumn{9}{c}{\textbf{Datasets: MSRS Fusion Dataset}}\\
     &EN&SD&SF&MI&SCD&VIF&Qabf&SSIM\\
    \midrule
    SDN     & 5.40 & 19.14 & 9.14  & 1.75 & 1.10 & 0.55 & 0.39 & 0.38 \\
    GANMcC  & 5.85 & 28.54 & 6.18  & 2.38 & 1.46 & 0.60 & 0.30 & 0.30 \\
    U2F     & 5.61 & 26.80 & 9.39  & 1.98 & 1.36 & 0.59 & 0.45 & {\bfseries\textcolor{blue}{0.40}} \\
    TarD    & 6.35 & 35.49 & 9.89  & 2.65 & 1.48 & 0.67 & 0.43 & 0.35 \\
    DDFM    & 5.56 & 32.84 & 8.43  & 2.47 & {\bfseries\textcolor{blue}{1.58}} & 0.67 & 0.48 & 0.31 \\
    CDDFuse & 6.29 & 42.47 & {\bfseries\textcolor{blue}{11.18}} & 4.67 & 1.55 & {\bfseries\textcolor{blue}{0.98}} & {\bfseries\textcolor{blue}{0.64}} & 0.36 \\
    SwinF   & {\bfseries\textcolor{blue}{6.61}} & {\bfseries\textcolor{blue}{43.69}} & 10.02 & 4.71 & 1.51 & 0.97 & 0.63 & 0.37 \\
    \textbf{Ours}    & {\bfseries\textcolor{red}{6.67}} & {\bfseries\textcolor{red}{43.74}} & {\bfseries\textcolor{red}{11.35}} & {\bfseries\textcolor{red}{4.78}} & {\bfseries\textcolor{red}{1.67}} & {\bfseries\textcolor{red}{1.00}} & {\bfseries\textcolor{red}{0.66}} & {\bfseries\textcolor{red}{0.47}} \\
    \midrule
    \multicolumn{9}{c}{\textbf{Datasets: RoadScene Fusion Dataset}}\\
     &EN&SD&SF&MI&SCD&VIF&Qabf&SSIM\\
    \midrule
    SDN     & 7.35 & 50.02 & {\bfseries\textcolor{blue}{15.35}} & {\bfseries\textcolor{blue}{3.47}} & 1.65 & 0.61 & 0.50 & 0.45 \\
    GANMcC  & 7.13 & 48.41 & 10.64 & 2.90 & {\bfseries\textcolor{red}{1.72}} & 0.51 & 0.39 & 0.41 \\
    U2F     & 7.22 & 42.23 & 14.94 & 2.84 & 1.59 & 0.57 & {\bfseries\textcolor{blue}{0.51}} & 0.28\\
    TarD    & 7.23 & 45.69 & 12.17 & 3.44 & 1.39 & 0.55 & 0.43 & 0.43 \\
    DDFM    & 7.36 & 49.07 & 13.36 & 3.05 & 1.65 & 0.60 & 0.50 & {\bfseries\textcolor{blue}{0.47}} \\
    CDDFuse & {\bfseries\textcolor{red}{7.43}} & {\bfseries\textcolor{red}{52.31}} & 14.62 & 3.14 & {\bfseries\textcolor{blue}{1.71}} & {\bfseries\textcolor{blue}{0.62}} & 0.45 & 0.45 \\
    SwinF   & 6.92 & 46.68 & 12.80 & 3.35 & 1.61 & 0.59 & 0.44 & 0.46 \\
    \textbf{Ours}    & {\bfseries\textcolor{blue}{7.38}} & {\bfseries\textcolor{blue}{51.30}} & {\bfseries\textcolor{red}{15.68}} & {\bfseries\textcolor{red}{3.62}} & 1.62 & {\bfseries\textcolor{red}{0.66}} & {\bfseries\textcolor{red}{0.56}} & {\bfseries\textcolor{red}{0.49}} \\
    \midrule
    \multicolumn{9}{c}{\textbf{Datasets: $\mathbf{M^{3}FD}$ Fusion Dataset}}\\
     &EN&SD&SF&MI&SCD&VIF&Qabf&SSIM\\
    \midrule
    SDN     & 6.67 & {\bfseries\textcolor{red}{40.83}} & 14.85 & 3.37 & 1.65 & 0.60 & 0.52 & 0.46 \\
    GANMcC  & 6.65 & 39.02 & 10.75 & 2.88 & 1.72 & 0.56 & 0.37 & 0.41 \\
    U2F     & 6.67 & 38.48 & 14.70 & 2.75 & {\bfseries\textcolor{red}{1.82}} & 0.68 & 0.56 & 0.47 \\
    TarD    & 6.68 & 40.02 & 12.87 & 3.17 & 1.51 & 0.59 & 0.42 & 0.44 \\
    DDFM    & 6.52 & 35.56 & 11.08 & 2.88 & {\bfseries\textcolor{blue}{1.81}} & 0.66 & 0.54 & 0.45 \\
    CDDFuse & 6.66 & 39.96 & {\bfseries\textcolor{red}{17.26}} & 3.85 & 1.68 & {\bfseries\textcolor{red}{0.82}} & 0.59 & {\bfseries\textcolor{blue}{0.48}} \\
    SwinF   & {\bfseries\textcolor{blue}{6.72}} & 37.52 & 14.22 & {\bfseries\textcolor{blue}{4.10}} & 1.60 & 0.73 & {\bfseries\textcolor{blue}{0.60}} & 0.47 \\
    \textbf{Ours}    & {\bfseries\textcolor{red}{6.81}} & {\bfseries\textcolor{blue}{40.19}} & {\bfseries\textcolor{blue}{16.89}} & {\bfseries\textcolor{red}{4.26}} & 1.75 & {\bfseries\textcolor{blue}{0.76}} & {\bfseries\textcolor{red}{0.61}} & {\bfseries\textcolor{red}{0.48}} \\
  \bottomrule
\end{tabular}
\end{table}
\textbf{Metrics.} We use eight metrics to quantitatively measure the fusion results: entropy (EN), standard deviation (SD), mutual information (MI), sum of the correlations of differences (SCD), spatial frequency (SF), visual information fidelity (VIF), Qabf and structural similarity index measure (SSIM). Higher metrics imply that a fusion image is better. \\
\textbf{Implement details.} The experiments use two NVIDIA GeForce RTX 4090 GPUs. A batch size 12 is employed, and each fusion task undergoes 10,000 training steps. During each step, images from the training set are randomly cropped into patches of size 128 × 128, which are subsequently normalized to fall within the range [0, 1]. The parameters of MambaDFuse are optimized using the Adam optimizer, with the learning rate set to $2\times10^{-5}$. The loss function is similar to that used in SwinFusion.\\
\textbf{For RGB inputs.} We adopt the same processing approach as previous work. Firstly, the RGB images are converted to the YCbCr color space, where the Y channel represents the luminance channel, and the Cb and Cr channels represent chrominance. Only the Y channel is utilized for fusion. Subsequently, the fused Y, Cb, and Cr channels are converted back to the RGB color space through inverse mapping.\\
\subsection{Comparison with SOTA methods}

\begin{figure*}
    \centering
    \includegraphics[width=1\linewidth]{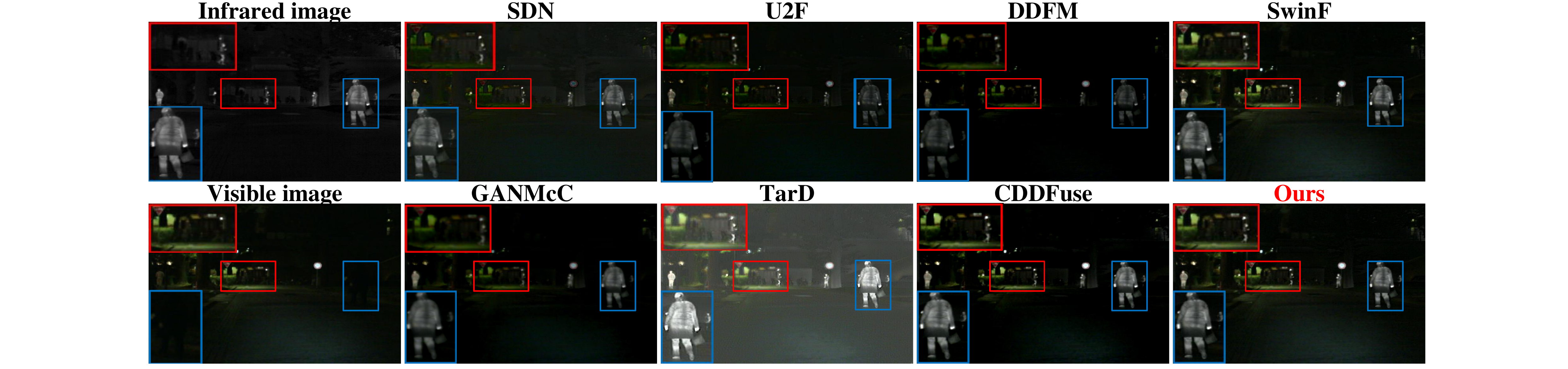}
    \caption{Visual comparison for “00718N” in MSRS IVF dataset.}
    \label{fig:msrs}
\end{figure*}

\begin{figure*}
    \centering
    \includegraphics[width=1\linewidth]{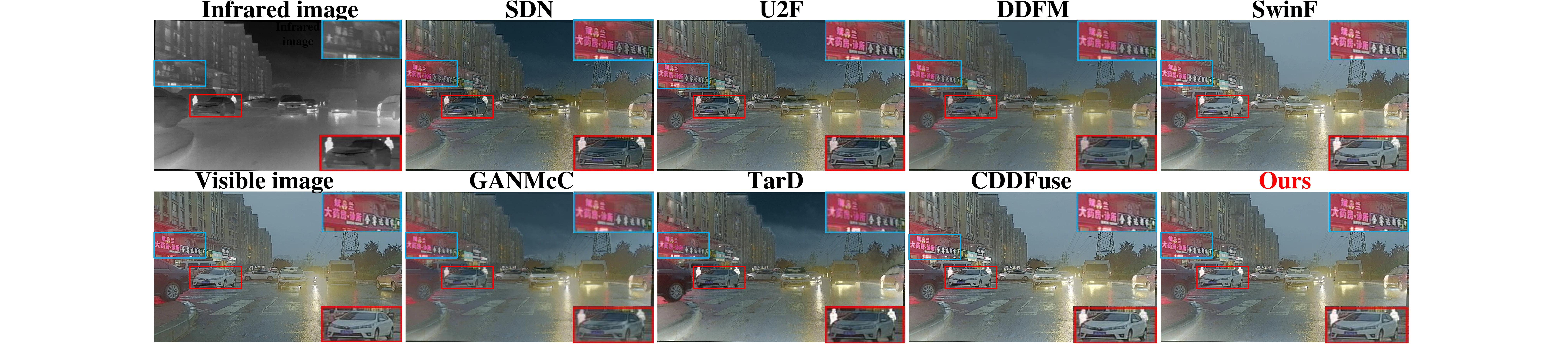}
    \caption{Visual comparison for “00033” in $\mathbf{M^{3}FD}$ IVF dataset.}
    \label{fig:m3fd}
\end{figure*}

\begin{figure}
    \centering
    \includegraphics[width=1\linewidth]{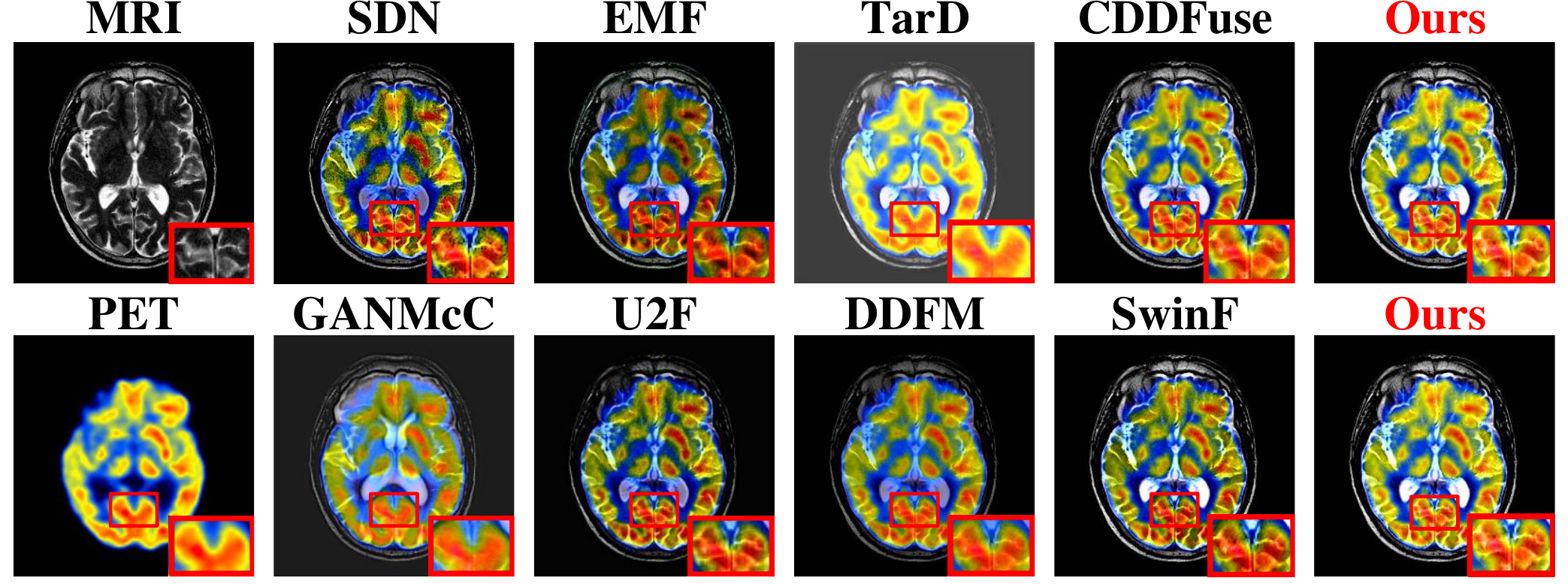}
    \caption{Visual comparison for MRI-PET MIF dataset.}
    \label{fig:mri-pet}
\end{figure}

\begin{table}\small
  \caption{Quantitative results of the MIF task. \bfseries\textcolor{red}{Red} and \bfseries\textcolor{blue}{blue} show the best and second-best values, respectively.}
  \label{tab:MIF}
  \begin{tabular}{ccccccccc}
    \toprule
    \multicolumn{9}{c}{\textbf{Datasets: MRI-CT Medical Fusion Dataset}}\\
     &EN&SD&SF&MI&SCD&VIF&Qabf&SSIM\\
    \midrule
    SDN     & 4.80 & 59.26 & 28.17 & 2.39 & 0.98 & 0.35 & 0.39 & 0.19  \\
    GANMcC  & 4.55 & 54.58 & 16.57 & 2.43 & 0.88 & 0.38 & 0.28 & 0.16  \\
    EMF     & 4.49 & 79.58 & 22.20 & 2.81 & 1.36 & 0.56 & 0.50 & 0.64  \\
    U2F     & 4.54 & 52.48 & 20.33 & 2.31 & 0.82 & 0.38 & 0.42 & 0.19  \\
    TarD    & {\bfseries\textcolor{red}{4.87}} & 62.41 & 17.86 & 2.65 & 0.85 & 0.45 & 0.34 & 0.17  \\
    DDFM    & 4.06 & 61.57 & 19.00 & {\bfseries\textcolor{red}{3.03}} & 1.12 & 0.47 & 0.40 & 0.62  \\
    CDDFuse & 4.37 & 84.01 & {\bfseries\textcolor{red}{33.81}} & 2.80 & 1.45 & 0.56 & {\bfseries\textcolor{blue}{0.60}} & {\bfseries\textcolor{blue}{0.66}}  \\
    SwinF   & 4.03 & {\bfseries\textcolor{blue}{87.14}} & 31.93 & 2.94 & \bfseries\textcolor{blue}{1.59} & {\bfseries\textcolor{red}{0.66}} & 0.57 & 0.65  \\
    \textbf{Ours}    & {\bfseries\textcolor{blue}{4.80}} & {\bfseries\textcolor{red}{90.95}} & {\bfseries\textcolor{blue}{32.43}} & {\bfseries\textcolor{blue}{2.99}} & {\bfseries\textcolor{red}{1.65}} & {\bfseries\textcolor{blue}{0.64}} & {\bfseries\textcolor{red}{0.60}} & {\bfseries\textcolor{red}{0.67}}  \\
    \midrule
    \multicolumn{9}{c}{\textbf{Datasets: MRI-PET Medical Fusion Dataset}}\\
     &EN&SD&SF&MI&SCD&VIF&Qabf&SSIM\\
    \midrule
    SDN     & 2.42 & 63.82 & {\bfseries\textcolor{blue}{30.44}} & 2.47 & 1.38 & 0.47 & 0.58 & {\bfseries\textcolor{red}{0.55}} \\
    GANMcC  & {\bfseries\textcolor{blue}{4.98}} & 58.64 & 16.29 & 2.23 & 0.92 & 0.29 & 0.24 & 0.45 \\
    EMF     & 4.73 & 60.91 & 25.69 & 2.79 & {\bfseries\textcolor{blue}{1.57}} & 0.58 & 0.58 & 0.52 \\
    U2F     & 4.61 & 60.46 & 22.54 & 1.15 & 1.21 & 0.40 & 0.49 & 0.47 \\
    TarD    & {\bfseries\textcolor{red}{5.01}} & 56.73 & 19.88 & 2.80 & 0.83 & 0.53 & 0.44 & 0.51 \\
    DDFM    & 4.39 & 63.79 & 18.95 & {\bfseries\textcolor{red}{3.64}} & 1.25 & 0.44 & 0.45 & 0.49 \\
    CDDFuse & 4.94 & 67.84 & 27.26 & {\bfseries\textcolor{blue}{3.23}} & 1.25 & {\bfseries\textcolor{blue}{0.64}} & 0.59 & 0.50 \\
    SwinF   & 3.90 & {\bfseries\textcolor{blue}{71.13}} & 30.33 & 2.93 & 1.50 & 0.63 & {\bfseries\textcolor{blue}{0.68}} & 0.51\\
    \textbf{Ours}    & 4.91 & {\bfseries\textcolor{red}{73.05}} & {\bfseries\textcolor{red}{30.45}} & 3.08 & {\bfseries\textcolor{red}{1.59}} & {\bfseries\textcolor{red}{0.65}} & {\bfseries\textcolor{red}{0.68}} & {\bfseries\textcolor{blue}{0.52}} \\
    \midrule
    \multicolumn{9}{c}{\textbf{Datasets: MRI-SPECT Medical Fusion Dataset}}\\
     &EN&SD&SF&MI&SCD&VIF&Qabf&SSIM\\
    \midrule
    SDN     & 3.80 & 44.74 & {\bfseries\textcolor{blue}{20.24}} & 2.40 & 1.27 & 0.58 & 0.64 & 0.62 \\
    GANMcC  & {\bfseries\textcolor{blue}{4.16}} & 44.61 & 11.41 & 1.86 & 1.10 & 0.31 & 0.21 & 0.36 \\
    EMF     & 3.98 & 42.86 & 15.09 & 2.40 & 0.88 & 0.66 & 0.67 & 0.59 \\
    U2F     & 3.98 & 50.01 & 15.74 & 1.97 & 1.09 & 0.39 & 0.47 & 0.58 \\
    TarD    & {\bfseries\textcolor{red}{4.25}} & 48.43 & 15.32 & 2.32 & 1.01 & 0.56 & 0.53 & 0.36 \\
    DDFM    & 3.89 & 43.57 & 12.92 & 2.99 & 1.05 & 0.49 & 0.44 & 0.56 \\
    CDDFuse & 4.09 & 52.12 & 18.10 & {\bfseries\textcolor{blue}{3.19}} & 1.19 & {\bfseries\textcolor{red}{0.82}} & 0.60 & 0.61 \\
    SwinF   & 3.90 & {\bfseries\textcolor{blue}{54.18}} & 20.10 & 3.03 & {\bfseries\textcolor{blue}{1.54}} & 0.81 & {\bfseries\textcolor{blue}{0.72}} & {\bfseries\textcolor{blue}{0.63}} \\
    \textbf{Ours}    & 3.99 & {\bfseries\textcolor{red}{55.43}} & {\bfseries\textcolor{red}{20.50}} & {\bfseries\textcolor{red}{3.32}} & {\bfseries\textcolor{red}{1.63}} & {\bfseries\textcolor{blue}{0.81}} & {\bfseries\textcolor{red}{0.75}} & {\bfseries\textcolor{red}{0.64}} \\
  \bottomrule
\end{tabular}
\end{table}

In this section, MambaDFuse is compared with the state-of-the-art methods, including the CNN and AE-based methods group: SDN \cite{zhang2021sdnet}, EMF \cite{xu2021emfusion}, U2F \cite{xu2020u2fusion}; and the generative methods group: GANMcC \cite{ma2020ganmcc}, TarD \cite{liu2022target}, DDFM \cite{zhao2023ddfm}; and the Transformer-based methods group: CDDFuse \cite{zhao2023cddfuse}, SwinF \cite{ma2022swinfusion}. Among them, EMF is an architecture designed specifically for the MIF task, so it is not compared in the IVF task.\\
\textbf{Quantitative comparison.} Eight metrics are employed to quantitatively compare the above results, which are displayed in Tab. \ref{tab:VIF} and \ref{tab:MIF}. MambaDFuse exhibited remarkable performance across nearly all metrics, confirming its applicability to various lighting conditions and object categories. In the results of our method, higher values of MI, Qabf, and SSIM indicate a higher amount of information transferred from the source images to the fused image with minimal distortion. Better SD suggests that MambaDFuse demonstrates improved contrast. Meanwhile, the excellent performance of SF also implies richer edge and texture details. The improved VIF further confirms that its ability aligns with human visual perception.\\
\textbf{Qualitative comparison.} Subsequently, we show the qualitative comparison in Fig. \ref{fig:msrs}, \ref{fig:m3fd} and \ref{fig:mri-pet}. MambaDFuse effectively integrates thermal radiation information from infrared images with detailed texture and illumination information from visible images. Consequently, objects located in dimly-lit environments are conspicuously accentuated, enabling easy distinguishing of foreground objects from the background. Moreover, previously indistinct background features due to low illumination now possess clearly defined edges and abundant contour information, enhancing their ability to comprehend the scene.

\subsection{Ablation Study}
\begin{figure}
    \centering
    \includegraphics[width=1\linewidth]{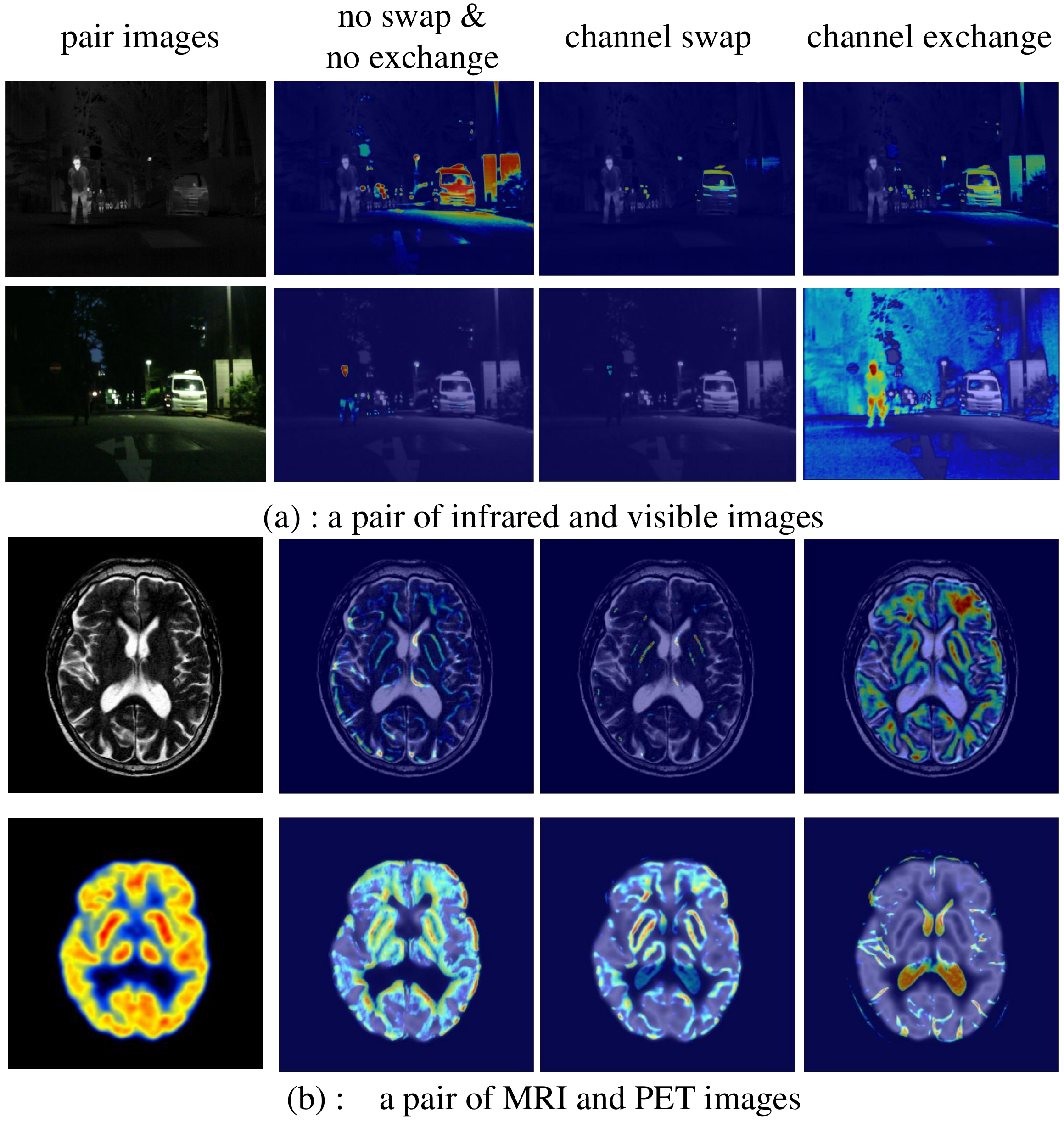}  
    \caption{Grad-CAM visualization results. Similar to Fig. \ref{fig:shallow} (b), we performed feature visualization on different options in the shallow fuse module and found that the features obtained using channel exchange excelled at integrating information from the other modality.}
    \label{fig:Ablation}
\end{figure}

\begin{table}\small
\centering
\caption{Ablation experiment results in the test set of MSRS. \bfseries\textcolor{red}{Red} indicates the best value.}
\label{tab:ablation_results}
\begin{tabular}{p{2.5cm}llllll}
\toprule
& EN & SD & SF & MI & VIF & Qabf \\
\midrule
\multicolumn{7}{l}{\textbf{Feature Extraction}} \\
\midrule
\quad I.w/o high-level feature extraction &6.65 &43.7 &11.24 &4.47 &0.97 &0.64 \\
\midrule
\multicolumn{7}{l}{\textbf{Shallow Fuse Module}} \\
\midrule
\quad II.w/o shallow fuse module& 6.66 &43.7 &11.29 &4.65 &0.98 &0.65 \\
\quad III.no swap \& no exchange& 6.66& 43.7&11.3 &4.67 & 0.99&0.65 \\
\quad IV.channel exchange $\rightarrow $ channel swap & 6.66&43.69 &11.34 & 4.75&1.00 &0.65 \\
\midrule
\multicolumn{7}{l}{\textbf{Deep Fuse Module}} \\
\midrule
\quad V.w/o deep fuse module&6.66 &43.72 &11.27 &4.59 &0.98 &0.64 \\
\quad VI.one modal as guidance& 6.65 &43.68 &11.21 &4.5 &0.97 &0.64 \\
\quad VII.another modal as guidance  & 6.65& 43.67& 11.24&4.57 &0.98 &0.64\\
\midrule
\textbf{Reconstruction} &  & & & & & \\
\midrule
\quad VIII.w/o Mamba blocks & 6.65& 43.68&11.26 &4.47 &0.97 &0.64\\
\midrule
\quad \textbf{Ours}  & {\bfseries\textcolor{red}{6.67}}&  {\bfseries\textcolor{red}{43.74}}& {\bfseries\textcolor{red}{11.35}} & {\bfseries\textcolor{red}{4.78}} & {\bfseries\textcolor{red}{1.00}} & {\bfseries\textcolor{red}{0.66}} \\
\bottomrule
\end{tabular}
\end{table}

Ablation experiments are conducted to verify the rationality of the different designs and modules. EN, SD, VIF, MI, VIF, and Qabf are utilized to validate the fusion effectiveness quantitatively. The results of the experimental groups are presented in Tab. \ref{tab:ablation_results}.

We performed ablation experiments across three stages: feature extraction, feature fusion, and fused image reconstruction. In the feature extraction stage, the effectiveness of high-level feature extraction was verified. In the feature fusion stage, experiments were performed by removing the shallow and deep fuse modules separately. The results showed a decrease in performance in both cases, demonstrating the necessity of both phases of feature fusion. Additionally, in the shallow fuse module, we investigated the options of using channel swap \cite{he2024pan}, channel exchange for feature interchange, or neither. Fig. \ref{fig:Ablation} visualizes the features, confirming the effectiveness of channel exchange in the module. In the deep fuse module, we ablated the selection of guiding in M3 block and found that the optimal metrics were achieved only when both modalities were used for guidance. Finally, in the fused image reconstruction, ablation experiments are conducted by removing the Mamba Blocks, confirming the design's rationale.

\subsection{Downstream IVF applications}

\begin{table}\scriptsize
\centering
\caption{The object detection performance (mAP) of visible, infrared, and fused images obtained from different methods on the MSRS dataset. \bfseries\textcolor{red}{Red} and \bfseries\textcolor{blue}{blue} show the best and second-best values, respectively.}
\label{tab:ap_values}
\begin{tabular}{cccc|ccc|ccc}
\toprule
 & \multicolumn{3}{c}{\textbf{AP@0.5}} & \multicolumn{3}{c}{\textbf{AP@0.7}} & \multicolumn{3}{c}{\textbf{AP@0.9}} \\
\cmidrule(lr){2-4} \cmidrule(lr){5-7} \cmidrule(lr){8-10}
 & Person & Car & All & Person & Car & All & Person & Car & All \\
\midrule
Infrared & 0.940 & 0.593 & 0.767 & 0.932 & 0.579 & 0.756 & \bfseries\textcolor{blue}{0.884} & 0.533 & 0.709 \\
Visible & 0.683 & 0.917 & 0.800 & 0.669 & 0.914 & 0.792 & 0.614 & 0.880 & 0.747 \\
SDN & 0.949 & 0.908 & 0.929 & {\bfseries\textcolor{red}{0.939}} & 0.911 & 0.92 & 0.863 & {\bfseries\textcolor{red}{0.904}} & 0.863 \\
GANMcC& 0.895 & {\bfseries\textcolor{red}{0.954}} & 0.919 & 0.889 & 0.870 & 0.909 & 0.789 & 0.825 & 0.852 \\
U2F & {\bfseries\textcolor{red}{0.951}} & 0.919 & 0.931 & 0.936 & 0.913 & {\bfseries\textcolor{red}{0.926}} & 0.854 & 0.876 & {\bfseries\textcolor{blue}{0.876}} \\
TarD & 0.929 & 0.927 & 0.928 & 0.927 & {\bfseries\textcolor{blue}{0.923}} & 0.921 & 0.859 & 0.888 & 0.873 \\
DDFM & 0.928 & 0.936 & {\bfseries\textcolor{blue}{0.932}} & 0.917 & {\bfseries\textcolor{red}{0.931}} & 0.924 & 0.837 & 0.890 & 0.863 \\
CDDFuse & 0.933 & 0.883 & 0.908 & 0.929 & 0.88 & 0.905 & 0.867 & 0.831 & 0.849 \\
SwinF & 0.896 & {\bfseries\textcolor{blue}{0.945}} & 0.92 & 0.883 & 0.891 & 0.912 & {0.873} & 0.863 & 0.868 \\
\textbf{Ours} & {\bfseries\textcolor{blue}{0.950}} & 0.940 & {\bfseries\textcolor{red}{0.935}} & {\bfseries\textcolor{blue}{0.937}} & 0.921 &{\bfseries\textcolor{blue}{0.925}} & {\bfseries\textcolor{red}{0.889}} & {\bfseries\textcolor{blue}{0.893}} & {\bfseries\textcolor{red}{0.879}} \\
\bottomrule
\end{tabular}
\end{table}

\begin{figure}
    \centering
    \includegraphics[width=1\linewidth]{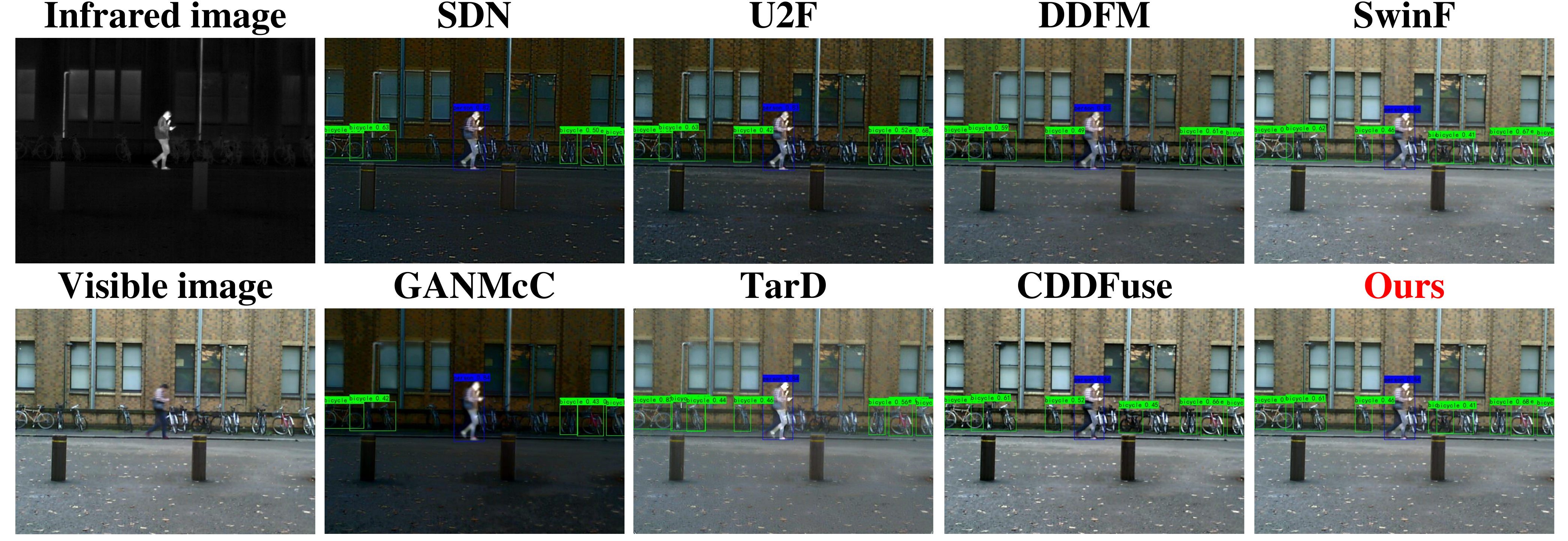}
    \caption{Results of object detection. The fused images obtained from MambaDFuse preserve both thermal radiation information and illumination information. Therefore, it makes objects in the image clearer and more distinguishable, achieving the best detection results.}
    \label{fig:object_det2}
\end{figure}

The purpose of MMIF is to facilitate further visual task applications, such as object detection, semantic segmentation, etc. Therefore, to determine the effectiveness of a method, the fused images should be utilized in downstream tasks to assess whether they positively contribute to these tasks. In our work, IVF image fusion is adopted for object detection. Eighty pairs of infrared and visible images were selected from the MSRS dataset, mainly annotated with people and cars. All fused images, together with the original infrared and visible images, were subjected to object detection using a pre-trained YOLOv5 model \cite{redmon2016you} (trained on the COCO dataset). Tab. \ref{tab:ap_values} presents the mean average precision (mAP), where AP@0.5, AP@0.7, and AP@0.9 denote the AP values at IoU thresholds of 0.5, 0.7, and 0.9, respectively. It was observed that visible images and infrared images provide limited information individually. For instance, object detectors are more likely to detect cars in visible images and people in infrared images. However, our fused images complement each other, providing a more comprehensive description of the scene and making it easier to detect both people and cars without any shortcomings. Among the various fusion methods compared, MambaDFuse demonstrates the best detection performance. Fig. \ref{fig:object_det2} also provides a visualized example.
\section{Conclusion}
In this paper, we explore Mamba's potential for multi-modality image fusion for the first time, proposing an efficient and effective Mamba-based dual-phase model and designing an Multi-modal Mamba (M3) block. On the one hand, dual-level feature extraction improves the extraction of modality-specific features. On the other hand, the dual-phase feature fusion module facilitates synthesizing comprehensive and complementary modality-fused features. Extensive experiments demonstrate that MambaDFuse can achieve promising fusion results and improve the accuracy of downstream tasks such as object detection.
\\
\\
\\
\\
\bibliographystyle{ACM-Reference-Format}
\bibliography{sample-base}










\end{document}